\documentclass[10pt,twocolumn,letterpaper]{article}

 \usepackage{cvpr}              

\usepackage{mathtools}
\usepackage{subcaption, caption}
\usepackage[sort&compress]{natbib}

\usepackage{booktabs}       
\usepackage{amsfonts}       
\usepackage{nicefrac}       
\usepackage{microtype}      

\usepackage{graphicx}
\usepackage{amsmath}
\usepackage{amssymb}

\usepackage{diagbox}
\usepackage{multicol}
\usepackage{enumerate}
\usepackage{times}
\usepackage{epsfig}
\usepackage{threeparttable}
\usepackage{enumitem}
\usepackage{multirow}
\usepackage{color}
\usepackage{array}
\usepackage{setspace}
\usepackage{makecell}
\definecolor{cvprblue}{rgb}{0.21,0.49,0.74}
\usepackage[pagebackref,breaklinks,colorlinks,allcolors=cvprblue]{hyperref}

\def\paperID{2940} 
\def\confName{CVPR}
\def\confYear{2025}

\title{TimeTracker: Event-based Continuous Point Tracking for Video Frame Interpolation with Non-linear Motion}

\author{Haoyue Liu\textsuperscript{1}, Jinghan Xu\textsuperscript{1},  Yi Chang\textsuperscript{1}\footnotemark[1], Hanyu Zhou\textsuperscript{1}, Haozhi Zhao\textsuperscript{1}, Lin Wang\textsuperscript{2}, Lunxin Yan\textsuperscript{1}\\
	\textsuperscript{1} National Key Lab of Multispectral Information Intelligent Processing Technology\\
	School of Artificial Intelligence and Automation, Huazhong University of Science and Technology\\
	\textsuperscript{2} School of Electrical and Electronic Engineering, Nanyang Technological University \\
	{\tt\small \{liuhy, xujinghan, yichang\}@hust.edu.cn}}

\begin{document}
\maketitle
\begin{abstract}
Video frame interpolation (VFI) that leverages the bio-inspired event cameras as guidance has recently shown better performance and memory efficiency than the frame-based methods, thanks to the event cameras' advantages, such as high temporal resolution. A hurdle for event-based VFI is how to effectively deal with non-linear motion, caused by the dynamic changes in motion direction and speed within the scene. Existing methods either use events to estimate sparse optical flow or fuse events with image features to estimate dense optical flow. Unfortunately, motion errors often degrade the VFI quality as the continuous motion cues from events do not align with the dense spatial information of images in the temporal dimension. In this paper, we find that object motion is continuous in space, tracking local regions over continuous time enables more accurate identification of spatiotemporal feature correlations. In light of this, we propose a novel continuous point tracking-based VFI framework, named \textbf{TimeTracker}. Specifically, we first design a Scene-Aware Region Segmentation (\textbf{SARS}) module to divide the scene into similar patches. Then, a Continuous Trajectory guided Motion Estimation (\textbf{CTME}) module is proposed to track the continuous motion trajectory of each patch through events. Finally, intermediate frames at any given time are generated through global motion optimization and frame refinement. Moreover, we collect a real-world dataset that features fast non-linear motion. Extensive experiments show that our method outperforms prior arts in both motion estimation and frame interpolation quality.
\end{abstract}

\section{Introduction}
\label{sec:intro}
High-frame-rate imaging is invaluable in scientific research, industrial inspection, security surveillance, and other fields. However, high-speed cameras produce massive data volumes, necessitating high-performance computing equipment for processing and storage, which adds to operational complexity and maintenance. VFI offers a cost-effective alternative for high frame rate imaging by inferring intermediate frames from spatiotemporal cues in neighboring frames, enabling the temporal upsampling of video frames.

Optical flow estimation \cite{pwc,raft} provides per-pixel displacement fields between frames, making it a widely used approach in VFI tasks \cite{superslomo,dain,softmax_splatting,bmbc,rife}. However, the loss of inter-frame information makes it challenging to capture the motion of the scene accurately. Existing methods typically assume linear motion between frames, but this assumption is often inadequate for real-world scenes with complex motion. To improve motion estimation accuracy, \cite{quadratic,chi2020all,zhang2020video,iqvfi} employs quadratic or cubic motion models, yet accurately capturing complex nonlinear motion between frames remains challenging. Inaccurate motion assumptions can lead to reconstruction artifacts, as shown in Fig.~\ref{fig2_motivation} (a1) and (b1).

\begin{figure}
	\setlength{\abovecaptionskip}{5pt}
	\setlength{\belowcaptionskip}{-12pt}
	\centering
	\includegraphics[width=0.99\linewidth]{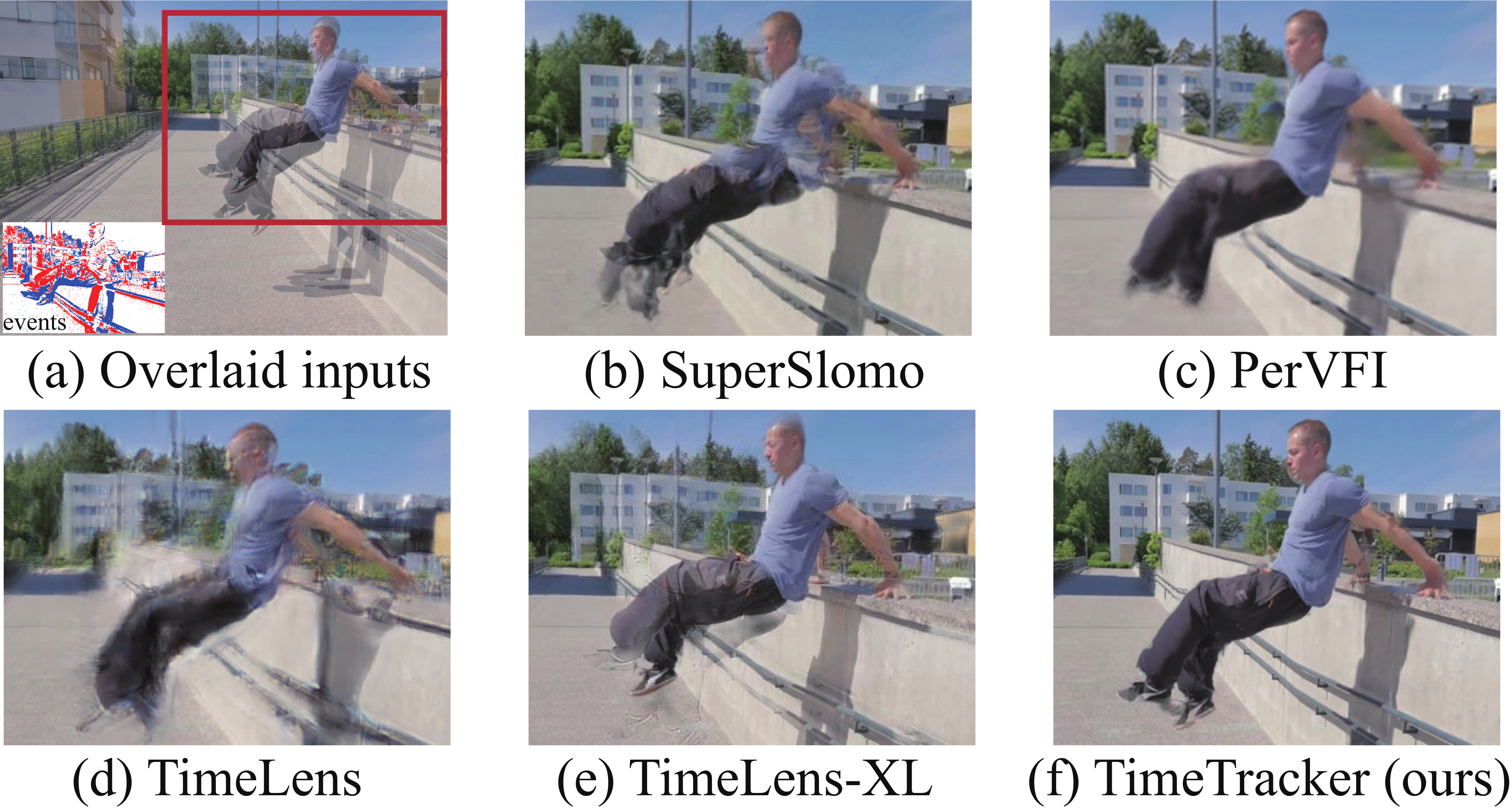}
	\caption{Visual comparison of our method with other SOTA methods. (b) and (c) estimate optical flow from images, (d) estimates optical flow from events, and (e) fuses image and event information to estimate optical flow. Our method, based on continuous point tracking for optical flow estimation, achieves the best performance.}
	\label{fig1_results}
\end{figure}

\begin{figure*}[t]
	\setlength{\abovecaptionskip}{5pt}
	\setlength{\belowcaptionskip}{-10pt}
	\centering
	\includegraphics[width=0.99\linewidth]{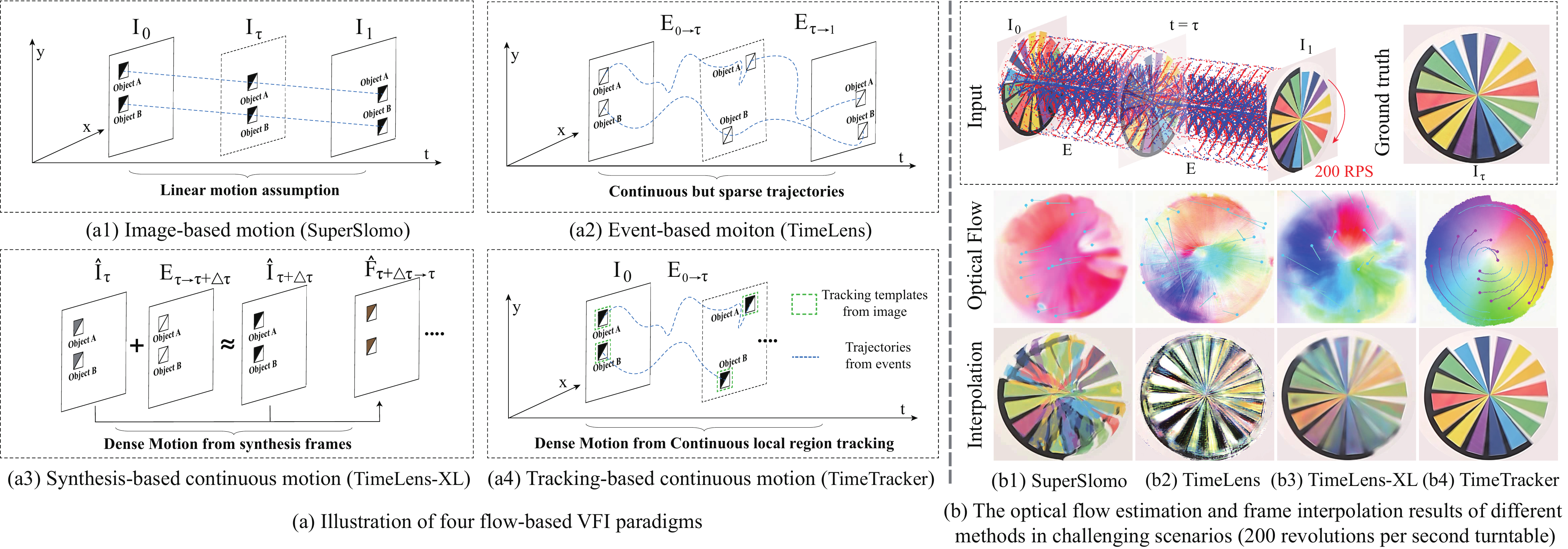}
	\caption{Illustration of (a) four flow-based VFI paradigms  and (b) their comparison results. Image-based methods like SuperSlomo \cite{superslomo} rely on a linear motion assumption, which results in significant inaccuracies in nonlinear motion scenarios. Timelens \cite{timelens} is a typical event-based VFI method that can only estimate sparse optical flow. TimelensXL \cite{timelens-xl} iteratively computes any-time optical flow by synthesizing intermediate frames, but errors in synthesized frames directly impact the accuracy of optical flow. We achieve dense any-time optical flow through image region segmentation and event-based point tracking, improving VFI performance in nonlinear scenarios.}
	\label{fig2_motivation}
\end{figure*}

Event cameras \cite{event_camera,event_survey} independently activate each pixel based on changes in brightness, capturing continuous motion edge information of objects. This characteristic offers several notable advantages, including high temporal resolution and lower bandwidth, making them a cost-effective data source for VFI. Consequently, event-based VFI presents an attractive approach. Synthesis-based methods \cite{eva2,superfast,e-vfi} directly fuse event and image features, with intermediate frames learned through a neural network. Flow-based methods \cite{timelens,timelens++,wevi,timereplayer,a2of,cbm,timelens-xl} estimate inter-frame optical flow directly from events \cite{timelens} or combine image-based and event-based optical flows \cite{timelens++, cbm} to determine the inter-frame optical flow. Compared to synthesis-based methods, flow-based approaches offer stronger physical constraints and greater robustness.

While event cameras can capture high temporal resolution motion information, their asynchronous triggering nature results in a highly sparse spatial distribution. Although accumulating events over longer periods can alleviate this sparsity, it inevitably reduces the temporal resolution of the event slices. Accumulating events for extended durations in fast and nonlinear motion scenes can introduce greater errors. Additionally, when the brightness change falls below the event trigger threshold (as in low-texture areas), no events are generated, further exacerbating spatial sparsity and making it insufficient for dense prediction tasks like VFI, as shown in Fig.~\ref{fig2_motivation} (a2) and (b2). Moreover, estimating high temporal resolution optical flow from reconstructed images creates a chicken-and-egg problem, where errors in the reconstructed images can accumulate in the optical flow estimation, ultimately reducing the accuracy of the final interpolation results, as shown in Fig.~\ref{fig2_motivation}  (a3) and (b3). The spatial sparsity of events poses a tricky challenge for event-based VFI: \textit{how can we obtain an accurate dense any-time optical flow from spatially dense but temporally discrete frames and spatially sparse but temporally continuous events?}

To address the above issues, we propose a novel VFI framework based on point tracking, TimeTracker. The core insight of the proposed method is to transform the any-time optical flow estimation problem into a local feature tracking problem. Events, which naturally capture high time resolution motion trajectories of fast-moving objects, help mitigate correlation-matching errors that typically arise during optical flow computation due to the sparsity of event data. This approach leverages the complementary nature of event and image modalities in the temporal and spatial dimensions, improving interpolation accuracy for high-speed and complex trajectory motion scenarios. 

Specifically, we first perform clustering and segmentation in the spatial domain based on the rich appearance features of the image. Locally similar regions in appearance tend to exhibit similar motion, especially for the rigid object. Next, we generate a motion region mask using the event trigger positions to distinguish dynamic regions from static ones. After sparsifying the image, we integrate the high temporal resolution motion trajectories provided by events for local feature tracking, resulting in a coarse dense any-time optical flow.  Since feature tracking is only performed within a limited spatial area, it avoids the errors caused by global feature matching in conventional optical flow estimation. Subsequently, a global attention module is used to optimize the optical flow iteratively, and the interpolation results for any moment are computed based on the refined optical flow. Finally, we use a frame optimization module to repair the regions with local optical flow estimation errors. The results of TimeTracker are shown in Fig.~\ref{fig1_results}, Fig.~\ref{fig2_motivation} (a4) and (b4). 

In addition, we build a coaxial imaging system and collect a challenging real-world paired Dataset of images and events featuring Complex, High-speed Motion (CHMD), which serves as an evaluation benchmark. Compare to existing datasets BS-ERGB \cite{timelens++}, ERF-X170FPS \cite{cbm} and HQ-EVFI \cite{timelens-xl}, CHMD features faster and more complex motion scenarios. It is carefully designed with controlled lighting conditions and optimized exposure times to minimize noise and motion blur. Overall, our main contributions can be summarized as follows:

\begin{itemize}[leftmargin=10pt]
	
	\item We propose the TimeTracker framework, which achieves any-time optical flow estimation through event-based point tracking. This method addresses the challenge of accurately estimating optical flow in high-speed and nonlinear motion scenes for VFI tasks, while also enabling multi-frame interpolation.
	
	\setlength{\itemsep}{5pt}
	\item We propose an any-time dense optical flow estimation strategy that fully leverages the advantages of both modalities in the temporal and spatial dimensions. By utilizing the rich appearance information from images to sparsify the scene in the spatial domain, we transform the any-time optical flow estimation problem into a local feature tracking problem, effectively avoiding motion estimation errors caused by the spatial sparsity of events.
	
	\setlength{\itemsep}{5pt}
	\item We introduce a challenging real-world paired dataset featuring complex, high-speed motion as an evaluation benchmark. Extensive experiments demonstrate that the proposed method achieves state-of-the-art performance across multiple datasets with complex motion.
	
\end{itemize}

\section{Related Work}
\label{sec:related_work}

\noindent
\textbf{Frame-based Video Frame Interpolation.}
Frame-based VFI has been widely studied and can generally be categorized into two main approaches: flow-based, and kernel-based methods. Flow-based methods \cite{superslomo,dain,softmax_splatting,bmbc,rife} explicitly estimate the motion between frames to generate intermediate latent frames. Kernel-based methods \cite{adaconv,sepconv,adacof} synthesize intermediate frames directly by applying convolution kernels within a network. Flow estimation \cite{pwc,raft} is widely used in VFI because it provides clear physical meaning and motion description. Although existing methods have shown promising results, challenges remain in complex motion scenarios due to missing inter-frame information. To improve motion estimation accuracy, some studies \cite{quadratic,chi2020all,zhang2020video,iqvfi} have introduced quadratic or cubic motion models , yet these approaches still struggle to model the intricate motion between frames.

\noindent
\textbf{Event-based Video Frame Interpolation.}
Event cameras \cite{event_camera,event_survey} provide high temporal resolution inter-frame visual information at a lower data rate, making event-based VFI a topic of growing interest in recent years \cite{timelens,timelens++,wevi,timereplayer,a2of,cbm,eva2,superfast,e-vfi,tta-evf,uniinr, timelens-xl}. Works like \cite{eva2,superfast,e-vfi} fuse high temporal resolution events and images directly to generate intermediate frames, however, the sparse nature of events can lead to artifacts in synthesized results. Timelens \cite{timelens} introduced a hybrid VFI framework that combines flow- and synthesis-based methods by estimating optical flow solely from events. Timelens++ \cite{timelens++}, A$^2$OF \cite{a2of} and CBM-Net \cite{cbm} estimates optical flow from both event and image features, however, the significant differences between the two modalities may lead to errors in the feature correlation calculation. The approach most similar to ours is TimeLens-XL \cite{timelens-xl}, which synthesizes intermediate frames using images and short-duration events, iteratively optimizing any-time optical flow and interpolation frames. However, its flow estimation depends on accurate frame synthesis, and errors in the synthesized frames can propagate as inaccurate optical flow over time.

In this work, we segment the image into similar local patches and track the motion trajectories of these patches, resulting in a non-linear and dense any-time optical flow. 

\noindent
\textbf{Continuous Motion Estimation.}
In recent years, frame-based point-tracking methods \cite{tap-vid,tapir,wang2023tracking,pointodyssey,cho2024local,cotracker} have made notable progress, allowing for the tracking of arbitrary points in images over extended time sequences. However, due to the low frame rate of images, these methods can only estimate optical flow over broader time scales, making them ineffective for capturing motion between frames. Event-based point-tracking methods \cite{eklt,messikommer2023data,fe-tap} offer high temporal resolution but produce only sparse tracking trajectories, which can be insufficient for dense motion estimation tasks.

B-Flow \cite{bflow} and MotionPriorCMax \cite{evimo2} attempt to densify event data by representing events as voxels, yet they struggle to produce accurate optical flow in sparse event regions. In contrast, Our approach fully leverages the rich appearance information from images to segment the scene into regions, initializes region-specific optical flow using point-tracking techniques, and refines this into dense, any-time optical flow using a global attention mechanism. This strategy effectively utilizes the spatial detail in images while avoiding the correlation errors that arise from attempting to estimate global motion from sparse event data alone.

\section{Event-based Dense Any-time Flow for VFI}
\label{sec3:method}

\begin{figure*}
	\setlength{\abovecaptionskip}{5pt}
	\setlength{\belowcaptionskip}{-10pt}
	\centering
	\includegraphics[width=0.99\linewidth]{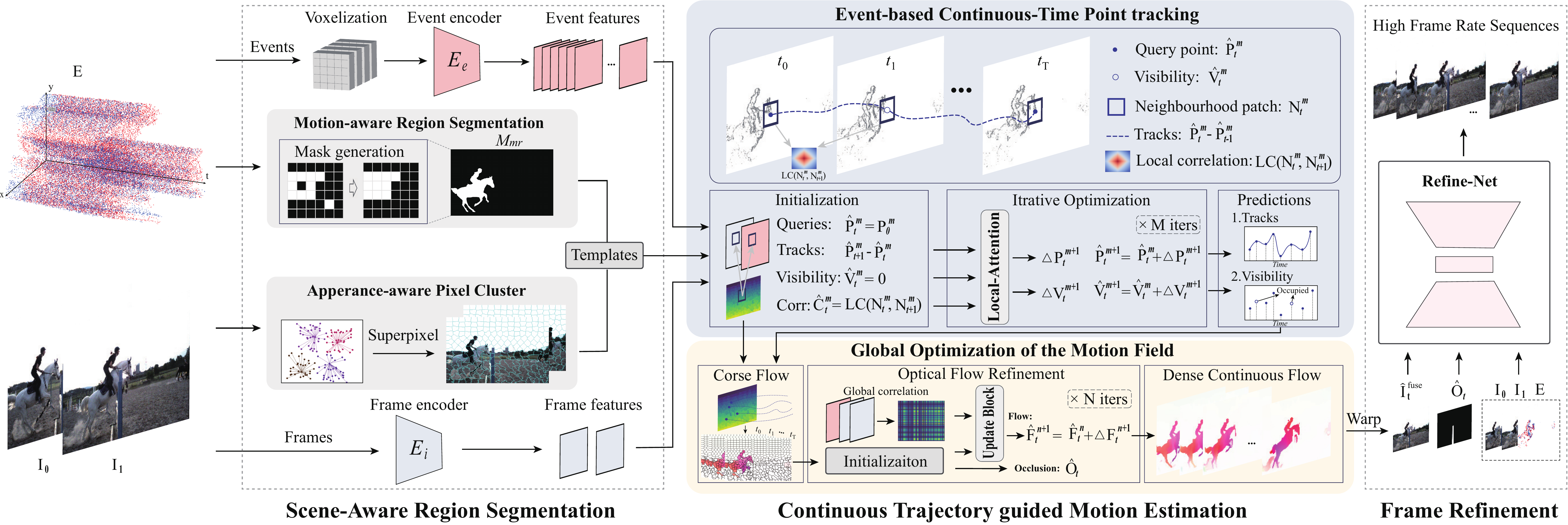}
	\caption{The overall architecture of the TimeTracker includes a Scene-Aware Region Segmentation (SARS) module, a Continuous Trajectory guided Motion Estimation (CTME) module, and a Frame Refinement (FR) module. The SARS segments the scene into multiple similar regions as tracking templates based on motion and appearance information. The CTME tracks the motion trajectories of each region and forms a dense any-time optical flow. Finally,  the FR refines the warped images to obtain the interpolation results.}
	\label{framework}
\end{figure*}

\subsection{Framework Overview}
\label{sec3.1:overview}
The key to achieving VFI in nonlinear motion scenes is accurately estimating dense, any-time optical flow between frames. Nonlinear and large displacement motion is decomposed into local linear and small displacement motion at high temporal resolution scales. To achieve this, we utilize the event modality to capture fine-grained temporal details. To address the sparsity of events in the spatial domain, we explore an optical flow densification strategy that combines the strengths of both image and event modalities. Specifically, we segment the image into regions with similar appearance and employ events for template-based region tracking. As a result, the dense optical flow estimation problem is transformed into a local feature tracking problem. Thanks to the inherent temporal continuity of motion, this approach is more robust compared to directly fusing the bimodal features. Finally, a frame optimization module is used to refine regions with local optical flow estimation errors, further enhancing the interpolation accuracy.  Fig.~\ref{framework} illustrates the overall framework of TimeTracker.

\subsection{Scene-Aware Region Segmentation}
\label{sec3.2:segmentation}
\noindent
\textbf{Motion-aware Region Segmentation.}
The event generation process \cite{event_survey} can be formulated as
\begin{equation}\
	\setlength\abovedisplayskip{3pt}
	\setlength\belowdisplayskip{3pt}
	\begin{aligned}
		\Delta L = logI(x, y, t) - logI(x, y, t - \Delta t) = p C,
		\label{eq:illu_dist}
	\end{aligned}
\end{equation}
where $logI(x, y, t)$ is the logarithmic illumination at pixel $(x, y)$ and time $t$, $\Delta t$ is the time interval between consecutive events, $p\in[-1, 1]$ is the polarity of events, and $C$ is the contrast threshold of the event camera. This process indicates that an event is triggered once the logarithmic illumination change at a particular pixel exceeds the threshold $C$. Assuming constant illumination \cite{event_survey}, for small $\Delta$t, equation (\ref{eq:illu_dist}) can be formulated as
\begin{equation}\
	\setlength\abovedisplayskip{3pt}
	\setlength\belowdisplayskip{3pt}
	\begin{aligned}
		\Delta L \approx - \nabla L \cdot v \Delta t,
		\label{eq:illu_dist_2}
	\end{aligned}
\end{equation}
where $\nabla L$ is the brightness gradient, $v$ is the optical flow. It can be observed that events are generated by the edges of object motion, while stationary regions with constant brightness generate no events. However, the differential imaging process is prone to generating isolated noise. To initialize a better tracking template, it is necessary to filter out regions with no events and those containing isolated noise from the region segmentation results. First, we represent all events $E=\left \{ e_i \right \} _{i=0}^{N-1}$ between boundary images $\left\{I_{0}, I_{1}\right\}$ as an event frame, ignoring the polarity of the events. Then, we apply morphological closing to connect broken regions in the event frame and remove isolated noise points, resulting in the motion region mask $M_{mr}$.

\noindent
\textbf{Appearance-aware Pixel Cluster.}
Directly inferring a dense optical flow from sparse events is an ill-posed problem. We aim to make full use of the rich appearance information in images by segmenting the scene into several small regions and then tracking motion trajectories within each region using the corresponding events. This approach enables us to initialize a coarse dense optical flow.

Rigid motion often exhibits spatial consistency, leading some methods \cite{seg_flow_cvpr2016,seg_flow_AAAI2020,seg_flow_ICCV2023} to leverage semantic information to enhance optical flow estimation by using distinctive object edges to maintain motion boundary accuracy. However, this assumption may break down in cases of dynamic textures, such as a waving arm, and directly using semantic segmentation models can incur high computational costs. Therefore, we adopt a simple yet effective method, SLIC \cite{slic}, to cluster pixels based on their intensity values and spatial positions, creating smaller segmented regions where motion consistency is easier to maintain at a finer spatial scale.

\begin{figure}
	\setlength{\abovecaptionskip}{5pt}
	\setlength{\belowcaptionskip}{-10pt}
	\centering
	\includegraphics[width=0.99\linewidth]{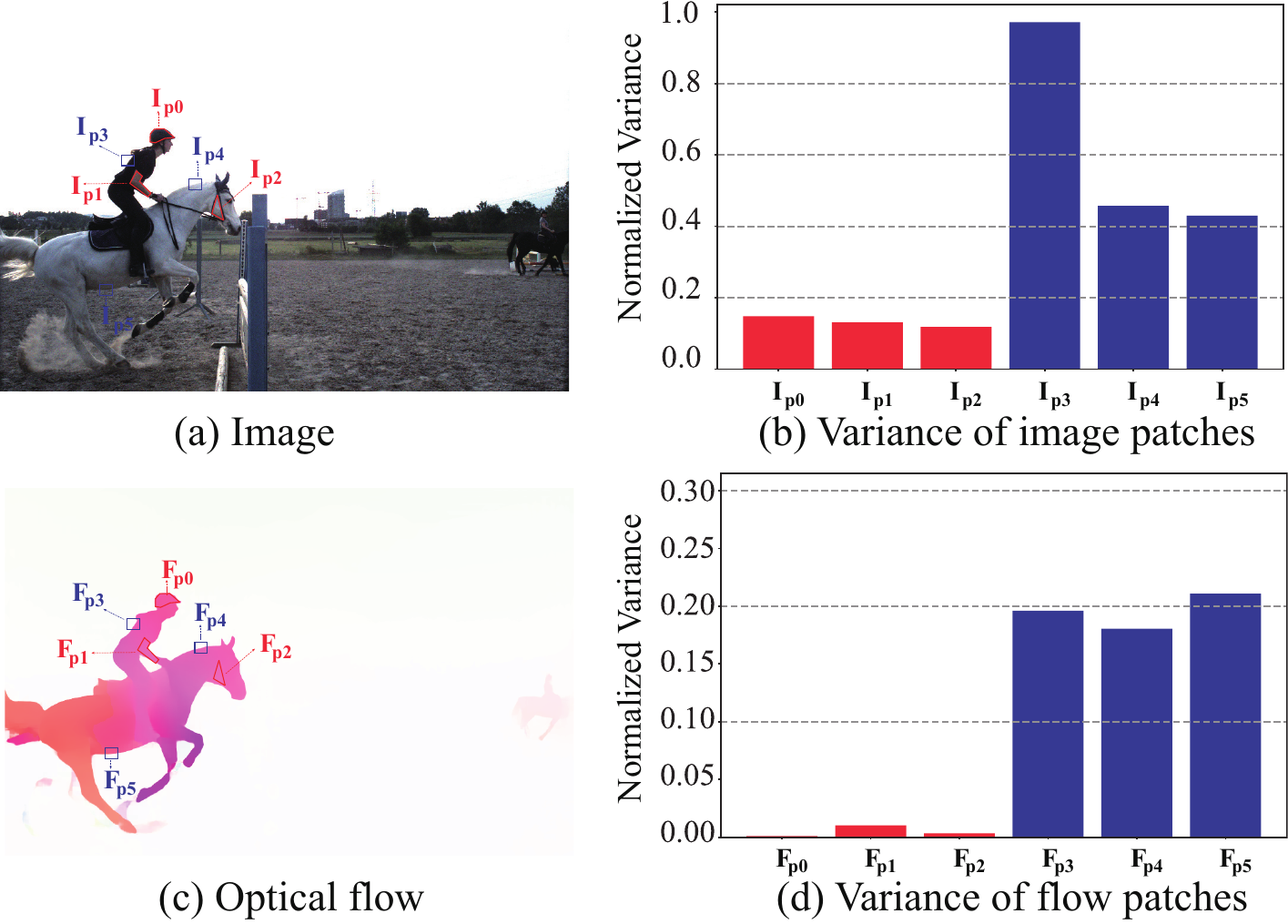}
	\caption{The correlation between appearance and motion. We select three regions with similar appearances, p0$\sim$p2 , in the image (a) and optical flow (b), along with three regions containing both foreground and background, p3$\sim$p5. (c) and (d) illustrate that small regions with minimal pixel value differences and spatial proximity in the image also exhibit consistent optical flow, and vice versa.}
	\label{fig4_motivation_seg}
\end{figure}

Fig.~\ref{fig4_motivation_seg} illustrates this concept: in the red box in Fig.~\ref{fig4_motivation_seg} (a), the moving foreground shares similar pixel values, while the blue box includes both foreground and background elements with more varied pixel values. Fig.~\ref{fig4_motivation_seg} (b) displays normalized pixel variance within each region, and Fig.~\ref{fig4_motivation_seg} (c) and Fig.~\ref{fig4_motivation_seg} (d) show the corresponding optical flow and normalized variance. It is evident that small regions with closely similar pixel values and spatial proximity exhibit consistent optical flow, and vice versa. Finally,  we use the motion mask $M$ to filter out invalid regions. Details and visual results of the SLIC can be found in the supplementary materials. 

\subsection{Continuous Trajectory Motion Estimation}
\label{sec3.3:motion_estimation}
\noindent
\textbf{Event Representation.}
The event stream needs be converted into tensor form for network input. We use voxels \cite{voxel} to represent the events. The voxelization method transform events $E_{k}=\left \{ e_i \right \} _{i=0}^{N-1}$ into a tensor $V\in \mathbb{R}^{B\times H \times W}$ with $B$ bins, which can be formulated as
\begin{equation}\
	\setlength\abovedisplayskip{3pt}
	\setlength\belowdisplayskip{3pt}
	\begin{aligned}
		\resizebox{0.8\hsize}{!}{$V(k)=\sum_ip_i\max(0, 1-\left |k- \frac{t_i-t_0}{t_N-t_0}(B-1) \right|)$},
		\label{eq:voxel}
	\end{aligned}
\end{equation}
where $N$ is the number of events, $p_i$ and $t_i$ represent the polarity and timestamp of the $i$-th event respectively, and the range of $k$ is in $[0, B - 1]$. 

\begin{figure}
	\setlength{\abovecaptionskip}{5pt}
	\setlength{\belowcaptionskip}{-10pt}
	\centering
	\includegraphics[width=0.99\linewidth]{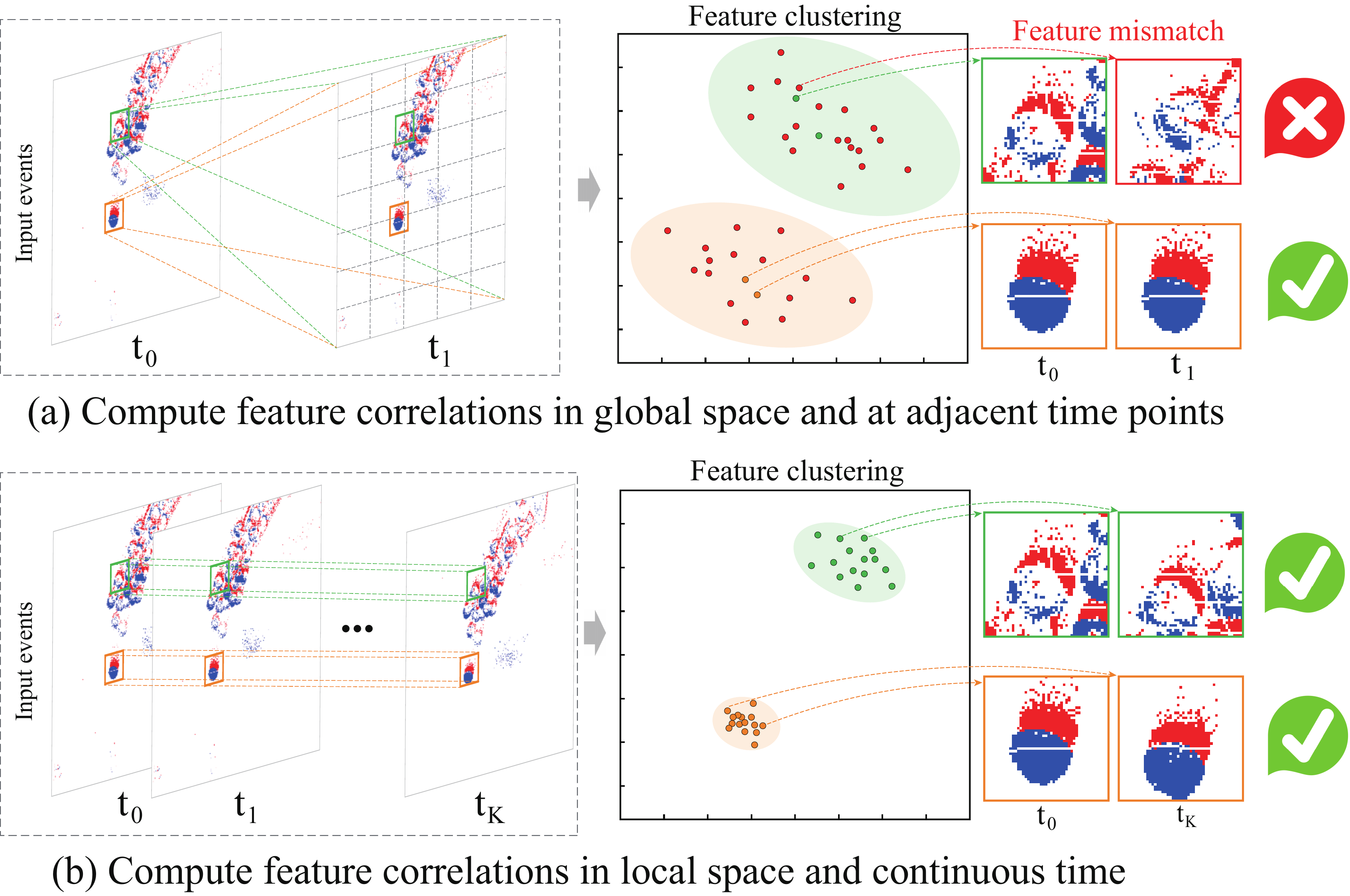}
	\caption{
		Two continuous motion estimation paradigms. (a) Traditional motion estimation methods calculate the cost volume between temporally adjacent features. Due to the low spatial distinguishability and sparsity of events, mismatches are likely to occur. (b) Since events are temporally continuous, performing continuous tracking within a local region is more robust.}
	\label{fig5_motivation_tracking}
\end{figure}

\noindent
\textbf{Event-based Continuous-Time Point Tracking.}
Compared to popular event-based optical flow methods like E-RAFT \cite{eraft}, which calculate the global cost volume between features at adjacent time points to obtain correlations,  we focus on tracking point trajectories over continuous time within a local region, and learn the displacement of the point over continuous time. Our motivation stems from the inherent property of events being triggered at motion edges, enabling them to capture continuous trajectories with high temporal resolution. However, events exhibit high spatial sparsity and distinguishability, which can lead to erroneous matches when computing the global cost volume of event features, as illustrated in Fig.~\ref{fig5_motivation_tracking}, the similarity of event features is higher in local space and continuous time (feature clustering is denser). Therefore, we select points from the regions filtered in Sec.~\ref{sec3.2:segmentation} for inter-frame tracking, which represent the motion trajectory of the region over continuous time.

Specifically, we begin by using SIFT \cite{sift} to identify easily trackable points on the boundary frames $\left\{I_{0}, I_{1}\right\}$. These points are then used to initialize a query point for each superpixel segment. If no event occurs at that location, we select the nearest event trigger location within the region as the query point. This process results in a set of query points.

Next, we use two separate feature extraction networks to extract multi-scale features from events and images, respectively. These features are generated across four down-sampling scales: $s \in \left\{ 4,8,16,32 \right\}$, with down-sampled features obtained via average pooling. Centered on each sample point, a square neighborhood $N_t$ is sampled at each scale using bilinear interpolation. In this region, the 4D correlation \cite{raft,cotracker} $C_t=LC(N_t, N_{t+1})$ between features at adjacent time points is calculated to track the position of the query point over time. $LC(\cdot)$  is obtained by stacking the inner products of features across multiple scales. 

Similar to \cite{cotracker,cho2024local}, we use a transformer \cite{transformer} block to establish attention over local space in continuous time, enabling motion trajectory tracking and optimization. First, for $i$-th query point at each time step $t$, we define the input tokens as follows:
\begin{equation}\
	\setlength\abovedisplayskip{3pt}
	\setlength\belowdisplayskip{3pt}
	\begin{aligned}
		G_{t}^{i} =\begin{pmatrix} 
		P_{t+1}^{i}-\hat{P}_{t}^{i},N_t^i,\hat{V}_{t}^{i},C_{t}^{i},\eta(\hat{P}_{t}^{i}-\hat{P}_{1}^{i}))
	 \end{pmatrix},
		\label{eq:s}
	\end{aligned}
\end{equation}
where $\hat{P}_{t+1}^{i}-\hat{P}_{t}^{i}$ is the predicted pixel displacement, $N_t^i$ is the content feature patch around query point, $\hat{V}_{t}^{i}$ is the predicted visibility, $C_{t}^{i}$ is the correlation volume, and $\eta(\cdot)$ denotes a sinusoidal positional encoding function. Then, we apply a sliding window of length $L$ between the boundary frames, moving forward by $L/2$ at each step. This can simultaneously model both the temporal correlation of a point trajectory over continuous time and the spatial correlation between different points. Each window undergoes $M$ optimization steps to yield the final tracks and visibility.

\noindent
\textbf{Global Optimization of the Optical Flow.}
Through point tracking, we can obtain the trajectory of each superpixel region over continuous time, resulting in a coarse, dense any-time optical flow $\hat{F}_{0 \to 1}^{coa}$ and a corresponding visibility mask $\hat{O}_{0 \to 1}$. Since potential errors may exist in point tracking, we further perform global optimization on the optical flow results. Our approach is based on the idea that optical flow and frame interpolation tasks should mutually reinforce each other, with interpolation results providing backward constraints to improve the accuracy of optical flow estimation.  Specifically, at high frame rates, large displacements and nonlinear motion can be approximated as locally linear motion. If the interpolation results are accurate, existing optical flow estimation methods \cite{raft,flowformer} should be able to produce relatively accurate optical flow. After excluding occluded regions, this optical flow should remain consistent with $\hat{F}_{0 \to 1}^{coa}$. Consequently, we construct a self-supervised consistency loss between them during training and update the parameters using a global optical flow optimization module, such as RAFT \cite{raft}, and iteratively optimize $N$ times during training. The global optimization module can simultaneously optimize both the optical flow and the occlusion. The loss function is defined as
\begin{equation}\
	\setlength\abovedisplayskip{3pt}
	\setlength\belowdisplayskip{3pt}
	\begin{aligned}
		\mathcal{L}_{flow}=\left \| (\hat{F}_{t-1\to t}^{coa}-\upsilon(\hat{I}_{t-1}, \hat{I}_{t}) ) \odot \hat{O}_{t} \right \| _1,
		\label{eq:flow_loss}
	\end{aligned}
\end{equation}
where $v(\cdot)$ denotes a frame-based optical flow estimation network, $\hat{I}_{t-1}$ and $\hat{I}_{t}$ are the VFI results, $\hat{O}_{t}$ is occlusion mask, and $\odot$ denotes element-wise multiplication. After global optimization, we obtain the refined dense any-time optical flow $\hat{F}_{0 \to 1}^{refine}$. In addition, we reverse the event stream following the method in \cite{timelens} and compute the backward optical flow $\hat{F}_{1 \to 0}^{refine}$ in the same way, thereby obtaining bidirectional any-time optical flow.

\subsection{Frame Interpolation and Refinement}
\label{sec3.4:frame_refine}
After obtaining the bidirectional optical flow, we warp the boundary images $\left\{I_{0}, I_{1}\right\}$ and fuse them using the method described in \cite{unflow,ocai} to generate the intermediate frame:
\begin{equation}\
	\setlength\abovedisplayskip{3pt}
	\setlength\belowdisplayskip{3pt}
	\begin{aligned}
		\resizebox{0.89\hsize}{!}{$\hat{I}_t^{fuse}=\frac{C_{t,0}}{C_{t,0}+C_{t,1}} w_b(I_0, \hat{F}_{t \to 0}^{refine}) + \frac{C_{t,1}}{C_{t,0}+C_{t,1}} w_b(I_1, \hat{F}_{t \to 1}^{refine})$},
		\label{eq:fuse_frame}
	\end{aligned}
\end{equation}
where $w_b(\cdot)$ denotes backward warping,  $\hat{F}_{t \to 0}^{refine}$ and $\hat{F}_{t \to 1}^{refine}$  are the intermediate flows sampled from the bidirectional any-time optical flow based on $t$, and $C_{0,1}$ is the confidence map, defined as follows:
\begin{equation}\
		\setlength\abovedisplayskip{3pt}
		\setlength\belowdisplayskip{3pt}
	\begin{aligned}
		\resizebox{0.89\hsize}{!}{$C_{0,1} = exp\left ( -\frac{\left | \hat{F}_{0 \to 1}(x) + \hat{F}_{1 \to 0}(x+\hat{F}_{0 \to 1}(x))  \right |^2} {\gamma _{1}(\left | \hat{F}_{0 \to 1} \right |^2+\left | \hat{F}_{1 \to 0}(x+\hat{F}_{0 \to 1}) \right | ^2 )+\gamma _{2}}  \right )$},
		\label{eq:flow_confidence}
	\end{aligned}
\end{equation}
where $\gamma _1=0.01$ and $\gamma _2=0.5$ from \cite{unflow}.
Due to the challenge of accurately estimating optical flow in occluded regions, we generate this part using a synthetic approach, which has been shown to be effective in previous work \cite{timelens,timelens++,cbm,e-vfi}. We first compute the occlusion regions $\hat{O}_{t}=\hat{O}_{t}^{forward}\cap \hat{O}_{t}^{backward}$, then input the fused image $\hat{I}_t^{fuse}$, boundary images $\left\{I_{0}, I_{1}\right\}$, events $E$, and occlusion mask $\hat{O}_{t}$ into the U-shape refine network. This network separately encodes the occlusion mask to provide attention for the refinement process. It then extracts information from events and boundary images to correct the erroneous regions in the fused image. Note that, after obtaining the any-time optical flow, we do not need to recursively interpolate frames at all time steps; instead, we perform selective interpolation based on the specified timestamp.

\subsection{Training Details}
\label{sec3.5:training_details}
\noindent
\textbf{Loss Function.}
The model loss function includes tracking loss, occlusion loss, reconstruction loss, and optical flow loss. The tracking loss and the occlusion loss are defined as:
\begin{equation}\
	\setlength\abovedisplayskip{3pt}
	\setlength\belowdisplayskip{3pt}
	\begin{aligned}
	\mathcal{L}_{track} =  {\textstyle \sum_{m=1}^{M}} \left \| \hat{P}^m_t - P^{GT}_t\right \| _1,
	\label{eq:loss_track}
	\end{aligned}
\end{equation}
\begin{equation}\
	\setlength\abovedisplayskip{3pt}
	\setlength\belowdisplayskip{3pt}
	\begin{aligned}
		\mathcal{L}_{occ} =  {\textstyle \sum_{m=1}^{M}} BCE(\hat{V}_t^m, V_t^{GT}),
		\label{eq:loss_occlusion}
	\end{aligned}
\end{equation}
where $BCE(\cdot)$ denotes binary cross entropy loss, $m$ is the number of iterations, $\hat{P}^m_t $ and $\hat{V}^m_t$ are the predicted point positions and occlusions, respectively.

The reconstruction loss is defined as:
\begin{equation}\
	\setlength\abovedisplayskip{3pt}
	\setlength\belowdisplayskip{3pt}
	\begin{aligned}
		\mathcal{L}_{rec} = \left \| \hat{I}_t - I_t^{GT} \right \| _1,
		\label{eq:loss_reconstrcution}
	\end{aligned}
\end{equation}
where $\hat{I}_t$ is the predicted frame.
The total loss during the training phase of the tracking model is:
\begin{equation}\
	\setlength\abovedisplayskip{3pt}
	\setlength\belowdisplayskip{3pt}
	\begin{aligned}
		\mathcal{L}_{total\_track} = \mathcal{L}_{track} + \lambda_1\mathcal{L}_{occ},
		\label{eq:loss_total_track}
	\end{aligned}
\end{equation}
the total loss during the training of the VFI model is:
\begin{equation}\
	\setlength\abovedisplayskip{3pt}
	\setlength\belowdisplayskip{3pt}
	\begin{aligned}
		\mathcal{L}_{total\_rec} = \lambda_2\mathcal{L}_{rec} + \lambda_3\mathcal{L}_{flow},
		\label{eq:loss_total_reconstruction}
	\end{aligned}
\end{equation}
we set $\lambda_1 = 1$, $\lambda_2 = 1$, and $\lambda_3 = 0.8$ respectively.

\noindent
\textbf{Implementation.}
We train the model in two steps. In the first step, we convert the Tap-Vid dataset \cite{tap-vid} to events using ESIM \cite{esim} and train the point tracking model on this dataset for 200K iterations. We then fine-tune on the Multiflow dataset \cite{bflow} for 15K iterations, using the ADAM \cite{adam} optimizer with a learning rate of 0.0005, the sliding window length is set to $L=10$, the iteration number is set to $M=5$. In the second step, we convert the GoPro dataset to events using ESIM \cite{esim}, freeze the tracking model, and train for 200K iterations with the ADAM \cite{adam} optimizer, starting with a learning rate of $10^{-4}$ and applying cosine decay down to $10^{-6}$. Training samples are cropped to 256×256, the iteration number is set to $N=10$.  All training is conducted using the PyTorch \cite{pytorch} on an NVIDIA A100 GPU.

\section{Experiments}
\label{sec:experiments}

\begin{figure*}
   \setlength{\abovecaptionskip}{5pt}
	\setlength{\belowcaptionskip}{-10pt}
	\centering
	\includegraphics[width=0.99\linewidth]{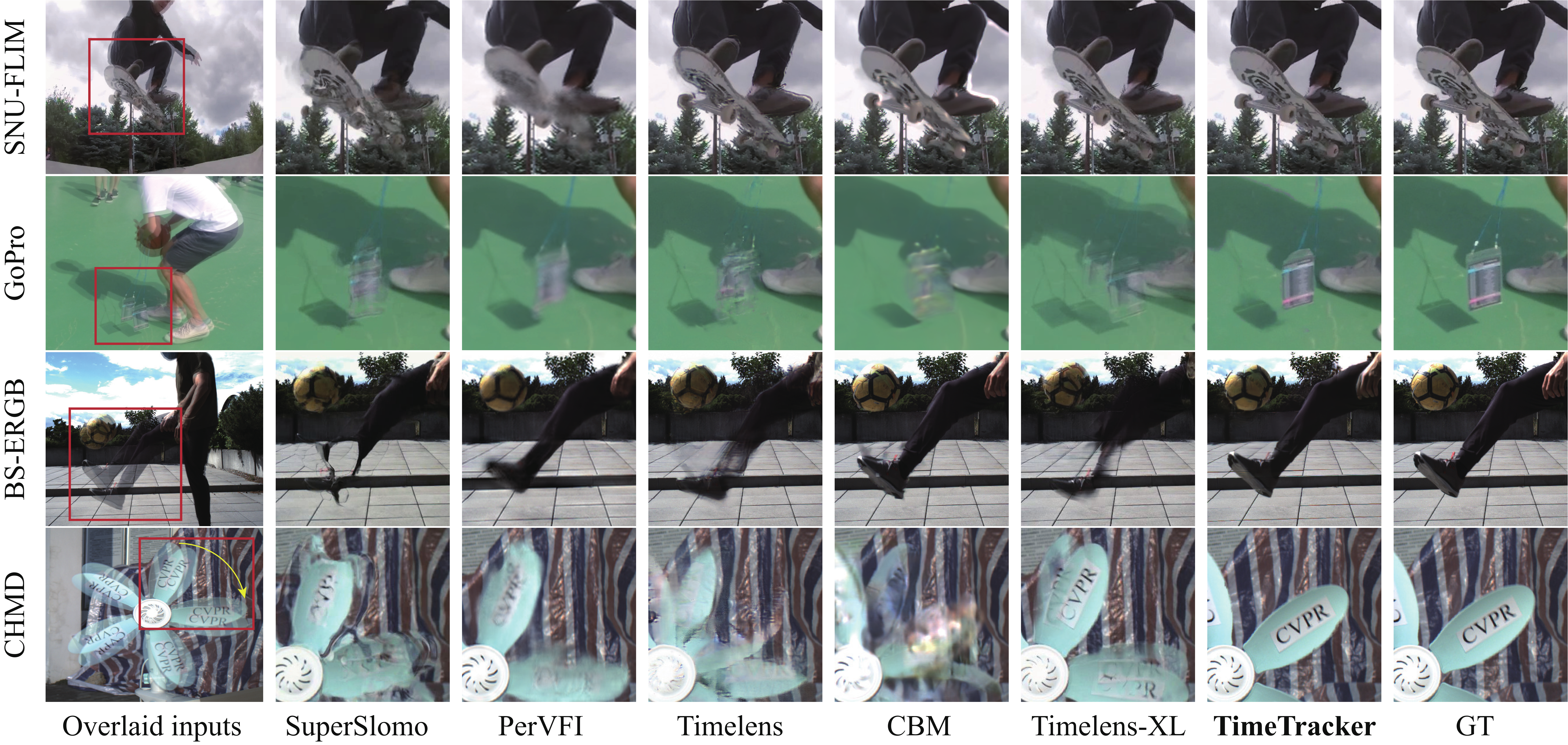}
	\caption{Visual comparison of the proposed method and other SOTA methods across different datasets.}
	\label{fig6_qualitive_evaluation}
\end{figure*}

\subsection{Datasets and Experimental Settings}
\noindent
\textbf{Datasets.}
Following the setup in previous works \cite{timelens,cbm,e-vfi,timelens-xl}, we evaluate the model on both synthetic and real event datasets using the same approach. For synthetic evaluation, we select GoPro \cite{gopro}, and SNU-FLIM \cite{snu_film}, generating events with ESIM \cite{esim}. For real-world data evaluation, we chose the BS-ERGB \cite{timelens++} and CHRD datasets, with CHRD including more challenging scenarios involving fast, nonlinear motion. Further details on CHRD are provided in the supplementary materials.

\noindent
\textbf{Comparison Methods.}
We compare our model with three frame-based SOTA methods: SuperSlomo \cite{superslomo}, PerVFI \cite{pervfi}, and VFIT-B \cite{vfit-b}, and four event-based SOTA methods: Timelens \cite{timelens}, CBMNet \cite{cbm}, SuperFast \cite{superfast}, and TimelensXL \cite{timelens-xl}. Among these, SuperSlomo \cite{superslomo} and PerVFI \cite{pervfi} are frame optical flow based  methods, Timelens \cite{timelens} is an event optical flow based method, CBMNet \cite{cbm} and TimelensXL \cite{timelens-xl} are event and image fusion optical flow methods, VFIT-B \cite{vfit-b} is an frame synthesis-based method, and SuperFast \cite{superfast} is a synthesis-based method that directly fuses images and events. Additionally, since Timelens \cite{timelens} and SuperFast \cite{superfast} were trained on different datasets, we fine-tune them on the GoPro dataset \cite{gopro} after loading their pretrained weights to ensure a fair comparison.

\subsection{Comparison Experiments}
\noindent
\textbf{Comparison on Synthetic Datasets.}
The quantitative and qualitative results on the synthetic datasets are reported in Tab.~\ref{tab1_quantitative_synthetic} and the first two rows of Fig.~\ref{fig6_qualitive_evaluation}. The proposed method outperforms existing methods in both PSNR and SSIM metrics. Frame-based methods, limited by the loss of inter-frame information, exhibit noticeable artifacts in areas with complex motion, and the reconstruction quality degrades significantly as more frames are skipped. Event-based methods, on the other hand, tend to show texture loss in dense-texture areas. In comparison, our method demonstrates a clear advantage in both visual quality and quantitative metrics. Benefiting from continuous trajectory tracking, TimeTracker achieves more stable results  when skipping more frames.

\begin{table}
	\footnotesize
	\renewcommand\arraystretch{1.1}
	\centering
	\setlength\tabcolsep{3pt}
	\setlength{\abovecaptionskip}{5pt}
	\setlength{\belowcaptionskip}{-12pt}
	\resizebox{0.5\textwidth}{!}{
		\begin{tabular}{ccccccccc}
			\Xhline{1px}
			\multirow{3}{*}{dataset} & \multicolumn{4}{c}{Gopro} & \multicolumn{4}{c}{SNU-FILM} \\ \cmidrule(r){2-5} \cmidrule(r){6-9}
			& \multicolumn{2}{c}{7skips} & \multicolumn{2}{c}{15skips} & \multicolumn{2}{c}{Hard} & \multicolumn{2}{c}{Extreme} \\ \cmidrule(r){2-3}\cmidrule(r){4-5}\cmidrule(r){6-7}\cmidrule(r){8-9}
			& PSNR         & SSIM        & PSNR         & SSIM         & PSNR         & SSIM        & PSNR         & SSIM         \\
			\hline
			SuperSlomo\cite{superslomo}  & 28.28 & 0.902 & 23.31 & 0.776 & 24.71 & 0.846 & 21.73 & 0.794        \\
			VFIT-B\cite{vfit-b}                   & 30.80  & 0.912 & 26.10 & 0.836 & 26.34  & 0.883 & 25.49 & 0.852        \\
			PerFVI\cite{pervfi}                   & 31.86  & 0.933 & 27.46 & 0.845 & 29.77  & 0.913 & 27.84  & 0.891        \\
			\hline
			Timelens\cite{timelens}         & 34.42 & 0.948 & 33.31 & 0.928 & 31.45 & 0.928  & 28.73 & 0.897        \\
			SuperFast\cite{superfast}       & 33.76 & 0.943 & 32.97 & 0.927 & 28.74  & 0.903  & 26.37 & 0.863        \\
			CBMNet\cite{cbm}                   & 36.86  & 0.955 & 35.32 & 0.947 & 30.87 & 0.918 & 27.56 & 0.884        \\
			Timelens-XL\cite{timelens-xl}  & 37.02 & 0.959 & 36.19 & 0.949 & 30.95 & 0.920 & 27.93  & 0.894        \\
			TimeTracker  (ours)              & \textbf{37.13}  & \textbf{0.962} & \textbf{36.54}  & \textbf{0.958} & \textbf{32.86} & \textbf{0.935} & \textbf{29.27} & \textbf{0.915}       
			\\	\Xhline{1px}  
	\end{tabular}}
	\caption{Quantitative results on synthetic datasets.}
	\label{tab1_quantitative_synthetic}
\end{table}

\noindent
\textbf{Comparison on Real Datasets.}
Tab.~\ref{tab2_quantitative_real} and the last two rows of Fig.~\ref{fig6_qualitive_evaluation} present the quantitative and qualitative results on real-world datasets. Event-based methods demonstrate better PSNR and SSIM metrics. On the CHRD dataset, which includes fast, nonlinear motion, frame-based method, frame-based methods incorrectly estimate the position of the fan blades, while other event-based methods exhibit severe artifacts. Our method demonstrates a significant advantage due to its accurate dense any-time optical flow.

\begin{table}
	\footnotesize
	\renewcommand\arraystretch{1.1}
	\centering
	\setlength\tabcolsep{3pt}
	\setlength{\abovecaptionskip}{5pt}
	\setlength{\belowcaptionskip}{-12pt}
		\resizebox{0.5\textwidth}{!}{
		\begin{tabular}{ccccccccc}
			\Xhline{1px}
			\multirow{3}{*}{dataset} & \multicolumn{4}{c}{BS-ERGB}                                & \multicolumn{4}{c}{Ours}                             \\ \cmidrule(r){2-5} \cmidrule(r){6-9}
			& \multicolumn{2}{c}{1skip} & \multicolumn{2}{c}{3skips} & \multicolumn{2}{c}{7skips} & \multicolumn{2}{c}{15skips} \\ \cmidrule(r){2-3}\cmidrule(r){4-5}\cmidrule(r){6-7}\cmidrule(r){8-9}
			& PSNR         & SSIM        & PSNR         & SSIM         & PSNR         & SSIM        & PSNR         & SSIM         \\
			\hline
			SuperSlomo\cite{superslomo}   & 23.33 & 0.734 & 22.43 & 0.716 & 21.17 & 0.705 & 20.26 & 0.674        \\
			VFIT-B\cite{vfit-b}                   & 24.44 & 0.741 & 24.31 & 0.725 & 22.04 & 0.743 & 21.70 & 0.682        \\
			PerFVI\cite{pervfi}                   & 27.72 & 0.761 & 26.07 & 0.763 & 24.82 & 0.768 & 21.65 & 0.702        \\
			\hline
			Timelens\cite{timelens}          & 28.13 & 0.787 & 26.82 & 0.769 & 25.86 & 0.771 & 24.12  & 0.748        \\
			SuperFast\cite{superfast}        & 27.87 & 0.768 & 26.77 & 0.758 & 22.79 & 0.762 & 20.59  & 0.722        \\
			CBMNet\cite{cbm}                  & 29.03 & 0.807  & 28.10  & 0.794  & 26.27  & 0.792 & 25.38  & 0.766        \\
			Timelens-XL\cite{timelens-xl} & 29.35 & 0.813 & 28.69 & 0.802  & 26.13  & 0.785 & 24.77  & 0.732        \\
			TimeTracker  (ours)                  & \textbf{29.85} & \textbf{0.823}  & \textbf{29.14}   & \textbf{0.807}   & \textbf{28.45}  & \textbf{0.814}   & \textbf{27.69}   & \textbf{0.805}       
			\\	\Xhline{1px}  
		\end{tabular}}
	\caption{Quantitative results on real-world datasets.}
	\label{tab2_quantitative_real}
\end{table}

\subsection{Ablation Study and Discussion}
\label{sec:ablation}

\begin{figure}
	\setlength{\abovecaptionskip}{5pt}
	\centering
	\includegraphics[width=0.95\linewidth]{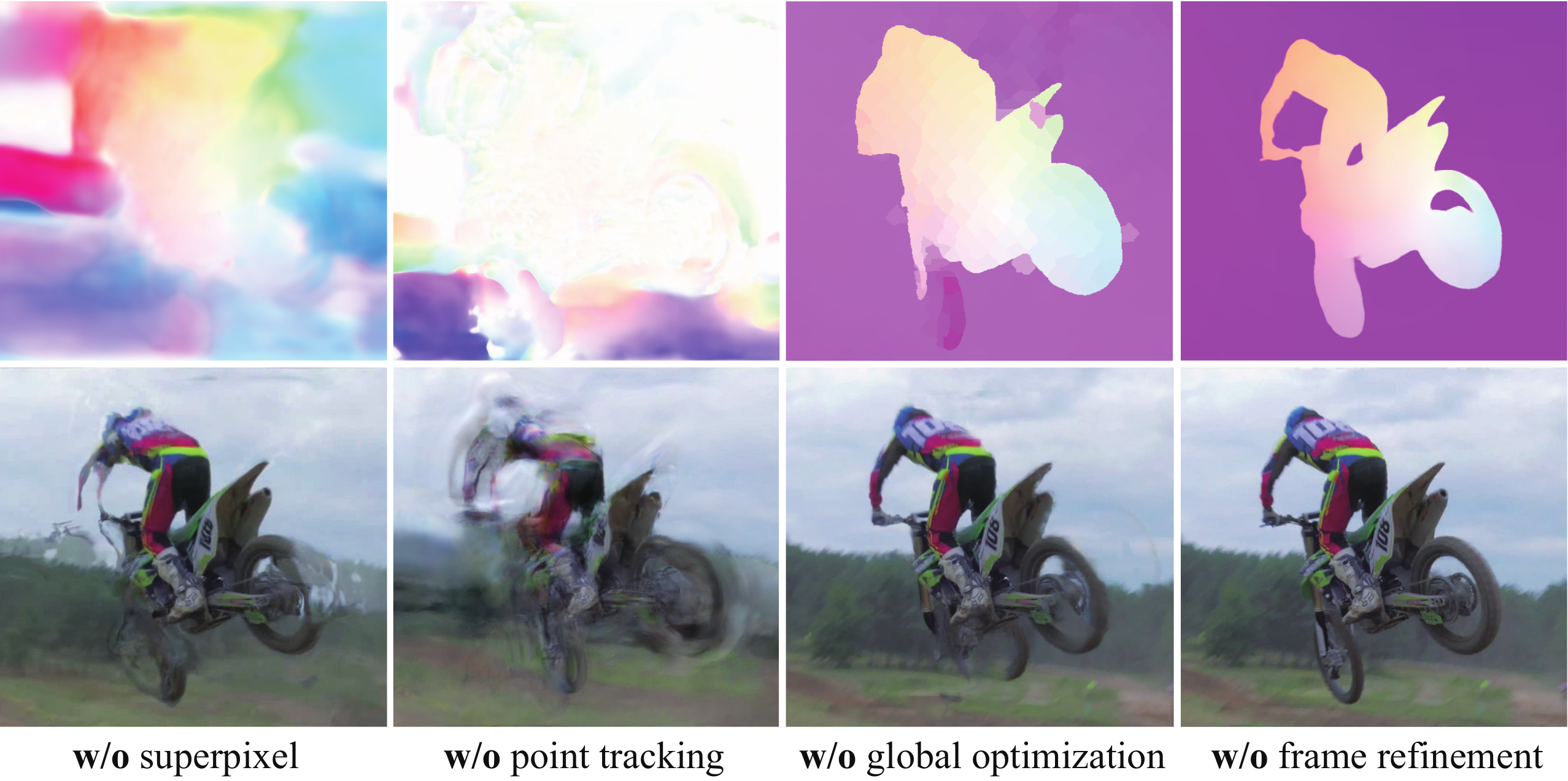}
	\caption{Ablation studies of TimeTracker. The first row and the second row show the optical flow and VFI results, respectively.}          
	\label{fig7_ablation_all}
\end{figure}

\noindent
\textbf{What role does each component of TimeTracker play?}
We study the roles of the main components in TimeTracker by removing them individually. Fig.~\ref{fig7_ablation_all} visually demonstrates the results of motion estimation and frame interpolation. We first replace the superpixel module with uniform image segmentation. The template obtained through uniform segmentation has pixels that do not exhibit motion consistency, and the incorrect initialization leads to poor motion global optimization results.  Removing the point tracking module results in zero-initialized optical flow, which fails to estimate the flow correctly. Removing the global optimization module causes artifacts from residual region segmentation in the optical flow. We further perform a quantitative analysis using the GoPro \cite{gopro} dataset, as shown in Tab.~\ref{tab:ablation_all}. The superpixel and point tracking modules are critical, while global optical flow optimization further improves reconstruction quality, with image refinement having a relatively weak effect.

\noindent
\textbf{Comparison of Motion Estimation Strategies.}
We further compare the effectiveness of the motion estimation modules in different VFI methods \cite{pervfi,timelens,cbm,timelens-xl}, as shown in Tab.~\ref{tab:ablation_point_tracking}. Specifically, we use the optical flow from the above methods to perform backward warping on the boundary frames and calculate the PSNR and SSIM metrics of the warped images. The tests are conducted on the GoPro dataset. PerVFI \cite{pervfi} exhibits optical flow errors due to its inaccurate motion prior assumptions. Event-based methods like TimeLens \cite{timelens} and CBMNet \cite{cbm} face limitations in motion estimation accuracy due to the inherent sparsity of events. While TimeLens-XL \cite{timelens-xl} attempts to address this by estimating optical flow through synthesized frames, which introduces additional instability. In contrast, TimeTracker achieves better results by utilizing point tracking-guided motion estimation.

\begin{table}
	\setlength{\abovecaptionskip}{5pt}
	\setlength{\belowcaptionskip}{-10pt}
	\footnotesize
	\centering
	\resizebox{0.485\textwidth}{!}{
		\begin{tabular}{cccc|cc}
			\Xhline{1px}
			Superpixel & Point tracking & Global optimization &  Frame refinement & PSNR~$\uparrow$& SSIM~$\uparrow$ \\
			\hline
			& \checkmark & \checkmark & \checkmark &30.62 & 0.925 \\
			\checkmark & &\checkmark &\checkmark  & 27.15 & 0.885 \\
			\checkmark & \checkmark & & \checkmark &32.43 & 0.933 \\
			\checkmark & \checkmark & \checkmark & &35.29 & 0.942 \\
			\checkmark & \checkmark & \checkmark & \checkmark & 37.47 & 0.965 \\
			\Xhline{1px}
	\end{tabular}}
	\caption{Ablation studies on main components of TimeTracker.}
	\label{tab:ablation_all}
	\vspace*{-2mm}
\end{table}

\begin{figure}
	\setlength{\abovecaptionskip}{5pt}
	\setlength{\belowcaptionskip}{-5pt}
	\centering
	\includegraphics[width=0.95\linewidth]{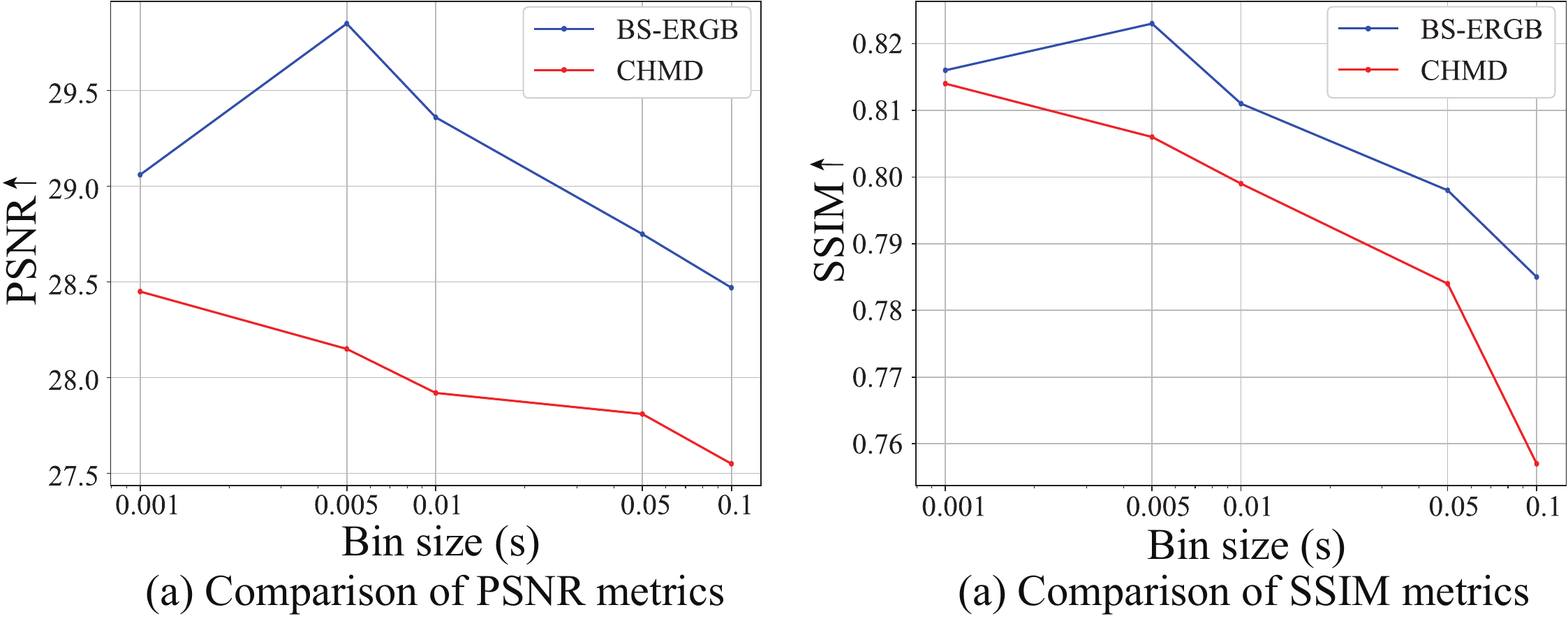}
	\caption{The impact of bin size settings on reconstruction quality.}
	\label{fig8_voxel_resolution}
	\vspace*{-0.5mm}
\end{figure}

\noindent
\textbf{Temporal Resolution of Voxel Grid.}
Events need to be voxelized before they can be converted into tensors that can be processed by the network. The shorter the time interval for generating the voxels, the smaller the displacement of objects within a unit voxel, leading to higher correlation between voxels at adjacent time steps, and vice versa. However, shorter time intervals also increase the computational load for point tracking within a fixed time period, and excessively short intervals may result in incomplete event features within a single voxel bin. In Fig~\ref{fig8_voxel_resolution}, we analyze the impact of different voxel bin sizes on VFI results in the BS-ERGB \cite{timelens++} and CHMD, where the objects in CHMD move relatively faster. It can be seen that in the BS-ERGB, the reconstruction quality is optimal when the voxel bin size is set to 0.005s, but in the CHMD, the reconstruction quality decreases as the voxel bin size increases. The reason is that objects with fast motion in CHMD have a higher event density, and high temporal resolution voxels help improve tracking performance.

\begin{table}
	\footnotesize
	\centering
	\resizebox{0.45\textwidth}{!}{
		\begin{tabular}{cccccc}
			\toprule
			Methods & PerVFI & TimeLens & CBMNet &  TimeLens-XL & TimeTraker\\
			\midrule
			Data source & {\makecell[c]{Image (I)}} & {\makecell[c]{Event (E)}} & {\makecell[c]{I+E}} & {\makecell[c]{I+E}} & {\makecell[c]{I+E}}\\
			\midrule
			PSNR~$\uparrow$ & 30.25 & 30.84 & 32.19 & 33.46  & \textbf{35.29} \\
			SSIM~$\uparrow$ & 0.908 & 0.916 & 0.925 & 0.939 & \textbf{0.942} \\
			\bottomrule
	\end{tabular}}
	\caption{Comparison of optical flow estimation performance.}
	\label{tab:ablation_point_tracking}
	\vspace*{-1mm}
\end{table}

\noindent
\textbf{Limitation and Future Work.}
Tracking-based motion estimation methods face limitations in dynamic texture scenes, such as fluids. The main challenge is that fluid features are temporally discontinuous (e.g., the shape of splashing water changes continuously, potentially appearing and disappearing rapidly). As a result, TimeTracker adopts an approach similar to existing methods \cite{timelens,timelens++,cbm}, directly synthesizing the relevant regions, which may lead to suboptimal results. However, event cameras inherently possess the ability to measure fluid motion. For instance, EBOS \cite{ebos} uses background-oriented schlieren method to measure gas flow fields, and EBIV \cite{ebiv} employs events to measure fluid particle velocities. In future work, we will further explore high frame rate imaging technology in dynamic textures scenes.

\section{Conclusion}
In this work, we propose TimeTracker, a novel point-tracking-based VFI framework that effectively adapts to complex nonlinear motion scenarios. To the best of our knowledge, this is the first study to address the VFI problem through continuous point tracking. We segment the scene into locally similar regions using the rich appearance features from the image, then track the continuous trajectories of these local regions using events, resulting in dense and any-time optical flow. Intermediate frames at any given time are generated through global motion optimization and frame refinement. Additionally, we introduce a dataset featuring fast nonlinear motion as a evaluation benchmark. The proposed method significantly outperforms state-of-the-art approaches, and we believe that our work can bring new perspectives to the community.

\clearpage
\newpage

\noindent
\textbf{Acknowledgments.}
This work was supported by the National Natural Science Foundation of China under Grant U24B20139 and 62371203, the Hubei Province Science Foundation of Distinguished Young Scholars under Grant JCZRJQ202500097, and the Start Up Grant at Nanyang Technological University under Grant 03INS002165C140. The computation is completed in the HPC Platform of Huazhong University of Science and Technology.

{
    \small
    \bibliographystyle{ieeenat_fullname}
    \bibliography{main}
}


\end{document}